\documentclass[conference]{IEEEtran}
\IEEEoverridecommandlockouts
\usepackage{graphicx}
\graphicspath{{./Visuals/}}
\usepackage{multirow}
\usepackage{tabularx}

\begin{document}
%
\title{Assessing Upper Limb Motor Function in the Immediate Post-Stroke Period Using Accelerometry}

\author{\IEEEauthorblockN{Mackenzie Wallich, Kenneth Lai, and Svetlana Yanushkevich}
\IEEEauthorblockA{Biometric Technologies Lab, Department of Electrical and Software Engineering, University of Calgary, Canada\\
\{mackenzie.wallich, kelai, syanshk\}@ucalgary.ca}}


%


\maketitle
	\IEEEpubid{
	\begin{minipage}{\textwidth}\ \\[70pt]
		\textbf{\footnotesize{{\fontfamily{ptm}\selectfont Digital Object Identifier 10.1109/CAI54212.2023.00064\\979-8-3503-3984-0/23/\$31.00 \copyright 2023 IEEE}}}
	\end{minipage}
}

\begin{abstract}
Accelerometry has been extensively studied as an objective means of measuring upper limb function in patients post-stroke. The objective of this paper is to determine whether the accelerometry-derived measurements frequently used in more long-term rehabilitation studies can also be used to monitor and rapidly detect sudden changes in upper limb motor function in more recently hospitalized stroke patients. Six binary classification models were created by training on variable data window times of paretic upper limb accelerometer feature data. The models were assessed on their effectiveness for differentiating new input data into two classes: severe or moderately severe motor function. The classification models yielded Area Under the Curve (AUC) scores that ranged from 0.72 to 0.82 for 15-minute data windows to 0.77 to 0.94 for 120-minute data windows. These results served as a preliminary assessment and a basis on which to further investigate the efficacy of using accelerometry and machine learning to alert healthcare professionals to rapid changes in motor function in the days immediately following a stroke.
\end{abstract}

\IEEEpeerreviewmaketitle

\section{Introduction}
Currently, professionals use standardized scoring systems to objectively assess both a patient’s neurological and physical state post-stroke \cite{teasell_salter_campbell_richardson_mehta_jutai_zettler_moses_mcclure_mays_et_al._2004}. A few examples of these scoring systems are the National Institutes of Health and Stroke Scale (NIHSS), the Action Research Arm Test (ARAT), and the Fugl-Meyer Assessment (FMA). These forms of testing can be time-consuming, and the results are prone to variability as they are repeated hourly and depend on the knowledge of the provider. Moreover, serious changes in patient well-being may go undetected between tests. Roughly $80\%$ \cite{PerssonHannaC2012Oaue} of stroke patients sustain upper limb impairment, where the behaviour of the affected upper limb is often characterized by weakness, spasticity, and loss of fine motor skills.

This paper aims to investigate accelerometry-derived measures and machine learning severity classification models as a potential starting place to develop a probabilistic doctor decision support application. 
In the present study, the validity of using six traditional binary machine learning models was tested by training the models using variable time windows of accelerometer feature data and outputting two classes: severe and moderately severe. The severe and moderately severe classes were differentiated using the ARAT score. The ARAT score was chosen to classify the two groups because it scores changes in upper limb function of stroke patients and has previously shown a moderate to strong correlation to accelerometry derived measures \cite{BarthJessica2020Cuem}. 

\section{Dataset}
The dataset used in this study was an accelerometry dataset from a prospective, longitudinal cohort of persons (n=67) with upper limb paresis post first time stroke \cite{LangCatherineE.2021ULPi}. The upper limb accelerometry data was collected from participants over a 24-hour period outside the clinic at 2, 4, 6, 8, 12, 16, 20, and 24 weeks post-stroke. Along with the accelerometer data, participants' demographics, affected arm, Action Research Arm Test (ARAT) scores, Upper Extremity Fugl-Meyer (UE-FM) scores, and Shoulder Abduction and Finger Extension (SAFE) scores were also recorded.

\section{Signal Pre-processing and Window Extraction}
Similar methods were replicated from \cite{8633876} for this section of the study. The data was sampled at 30Hz so no down sampling was required. The vector magnitude was calculated for each accelerometer data row; however, the raw accelerometer data was used as opposed to the activity counts. The use of 12.8s non-overlapping windows was proposed by \cite{SHAOYANZHANG2012PACU} for physical activity classification using wrist worn accelerometers. In \cite{8633876}, movement events were extracted as windows; however, the process to detect movement events proved difficult so 12.8s windowing was used instead. Finally, the first ten minutes of data was cut from the accelerometry data and only the following 4 hours were considered when generating the feature matrix. This was done so the ARAT score evaluated in the clinic before putting on the accelerometers remains relevant to the data collection period.
\begin{figure*}[hbt!]
	\centering
	\includegraphics[width=0.72\textwidth]{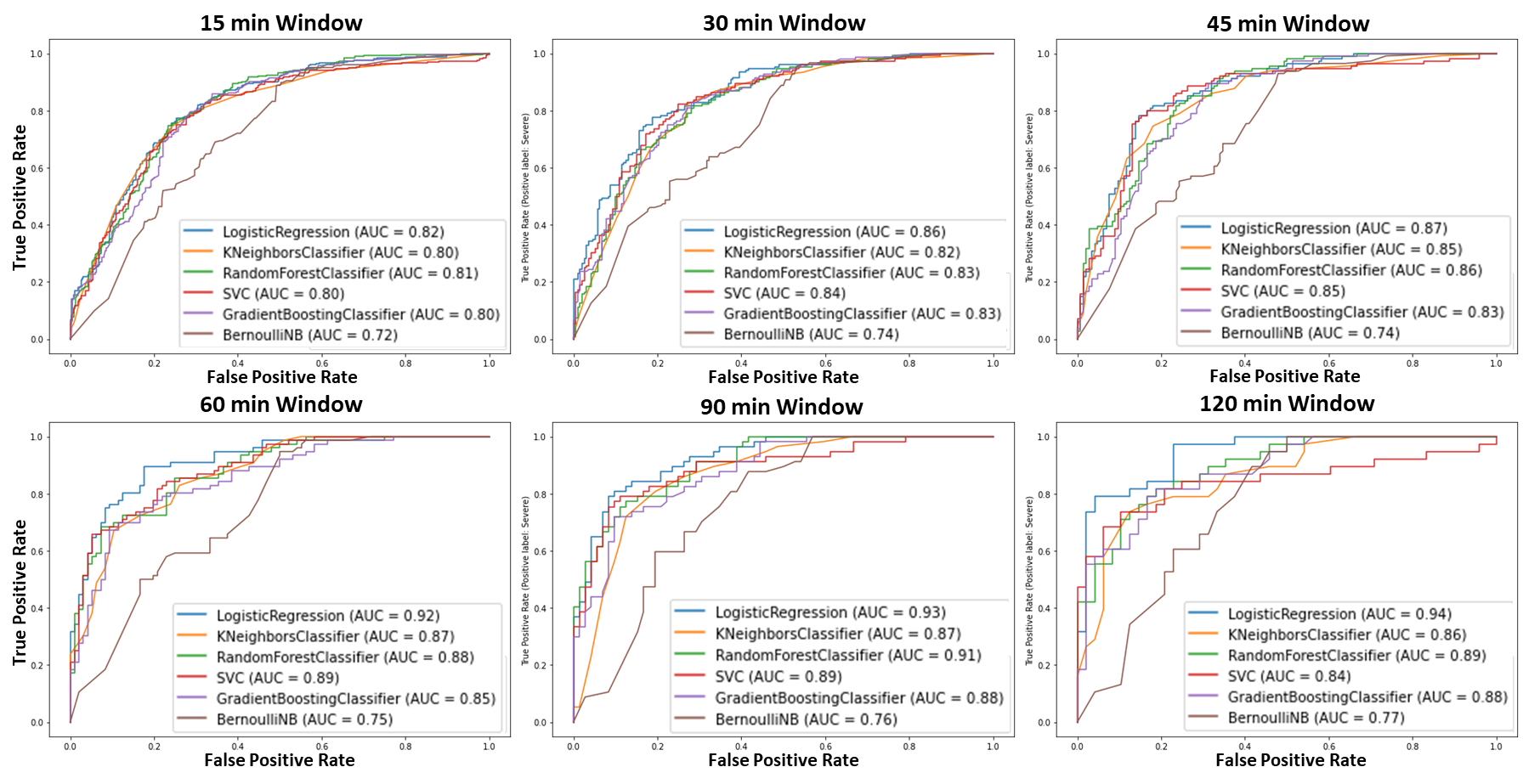}
	\caption{The ROC curves for each model grouped by window length}
	\label{figure:roc}
\end{figure*}

\section{Feature Generation}
The features selected for use in this study were taken directly from \cite{8633876}, with the exception of Movement Counts, as they were shown to yield high classification results. The features characterized smoothness, intensity, and pattern behaviour in both upper and lower limb movements. Features such as the magnitude average, maximum magnitude, minimum magnitude, NARJ, frequency 1, frequency 2, power 1, and power 2 were extracted from both the time and frequency domain \cite{SHAOYANZHANG2012PACU} and included in the feature matrix. The frequency domain features were extracted by doing an FFT (Fourier Fast Transform) of the vector magnitude then extracting the first and second dominant frequencies (excluding DC) and their corresponding powers. The features were extracted from each 12.8s window then the average of all the features for various window times were calculated resulting in one feature vector for each 15 min, 30 min, 45 min, 60 min, 90 min, or 120 min window. Combining the features this way minimizes the effects of movement generated by clinicians interacting with the patient and provides a more accurate feature for movement over the entire data collection period.

\section{Machine Learning Models}
Logistic Regression, Naive Bayes, Random Forest, Gradient Boosting, and K-Nearest Neighbors classifiers were used along with Support Vector Machines to solve this binary classification problem. These models were selected because they are popular for use in binary classification.

The two classes for the model were “severe” and “moderate” paretic limbs. The data was divided so approximately $80\%$ of the patients were in the training set and the other $20\%$ were in the test set. The sets were split by patients to ensure that the test data had never been seen before. A grid search with 5-fold cross validation was performed with each of the classifiers and window times to determine the best combination of hyperparameters. The model performance was evaluated using the area under the receiver operating characteristics curve (AUC-ROC). The curve plots the true positive rate versus the false positive rate. The AUC-ROC curve measures a model's performance at distinguishing between two classes.

\section{Severity Classification Results}

There was a strong Pearsons correlation between the ARAT score and use ratio (0.866). From this, it was inferred that ARAT would be an appropriate target for paretic limb severity. The results from the trained models and windows are show in Fig. \ref{figure:roc}. The Logistic Regression model had the highest AUC score of 0.94 for a window length of 120 minutes; however, this score was only marginally better than the AUC scores for 90 minutes and 60 minutes, which were 0.93 and 0.92, respectively.

\section{Conclusion}
The combination of accelerometry-derived measures from the literature and the visuals presented in this paper created a preliminary decision support tool for healthcare professionals. While this may prove to be an important tool for health care providers, clinical performance measures, such as ARAT scores, should remain an element of patient care as they provide additional information about paretic limb recovery as well as face-to-face contact with patients. This study also showed that in addition to the more popular Support Vector Machines, Logistic Regression should also be considered when creating a probabilistic machine learning severity classification model. Finally, the methods applied in this research may be used for other decision support tools in the future.


\section*{Acknowledgment}
%
This work was supported in part by the Undergraduate Research Experience (PURE) at the University of Calgary.



%
\bibliographystyle{IEEEtran}
\bibliography{citation}

\end{document}